\documentclass{article}

\usepackage{microtype}
\usepackage{graphicx}
\usepackage{subfigure}
\usepackage{booktabs}
\usepackage[table]{xcolor}
\usepackage{hyperref}
\usepackage{multirow}
\usepackage{xcolor}

\usepackage[accepted]{icml2025}

\usepackage{amsmath}
\usepackage{amssymb}
\usepackage{mathtools}
\usepackage{amsthm}

\usepackage[capitalize,noabbrev]{cleveref}

\theoremstyle{plain}

\theoremstyle{definition}

\theoremstyle{remark}

\icmltitlerunning{Understanding When GCN Helps: A Study on Label Scarcity and Graph Structure}

\begin{document}

\twocolumn[
\icmltitle{Understanding When Graph Convolutional Networks Help: \\
           A Diagnostic Study on Label Scarcity and Structural Properties}

\icmlsetsymbol{equal}{*}

\begin{icmlauthorlist}
\icmlauthor{Nischal Subedi}{udel}
\icmlauthor{Ember Kerstetter}{udel}
\icmlauthor{Winnie Li}{udel}
\icmlauthor{Silo Murphy}{udel}
\end{icmlauthorlist}

\icmlaffiliation{udel}{Department of Applied Economics and Statistics, University of Delaware, Newark, DE, USA}

\icmlkeywords{Graph Neural Networks, Semi-supervised Learning, Node Classification, Label Scarcity}

\vskip 0.3in
]

\begin{abstract}
Graph Convolutional Networks (GCNs) have become a standard approach for semi-supervised node classification, yet practitioners lack clear guidance on when GCNs provide meaningful improvements over simpler baselines. We present a diagnostic study using the Amazon Computers co-purchase data to understand when and why GCNs help. Through systematic experiments with simulated label scarcity, feature ablation, and per-class analysis, we find that GCN performance depends critically on the interaction between graph homophily and feature quality. GCNs provide the largest gains under extreme label scarcity, where they leverage neighborhood structure to compensate for limited supervision. Surprisingly, GCNs can match their original performance even when node features are replaced with random noise, suggesting that structure alone carries sufficient signal on highly homophilous graphs. However, GCNs hurt performance when homophily is low and features are already strong, as noisy neighbors corrupt good predictions. Our quadrant analysis reveals that GCNs help in three of four conditions and only hurt when low homophily meets strong features. These findings offer practical guidance for practitioners deciding whether to adopt graph-based methods.
\end{abstract}

\section{Introduction}\label{sec:introduction}

Labeling data is expensive. In many real-world applications, from e-commerce product categorization to social network analysis, only a small fraction of data points come with ground-truth labels. This label scarcity poses a fundamental challenge for supervised learning methods that rely on large labeled datasets to generalize well.

Graph Convolutional Networks (GCNs) have emerged as a promising solution to this challenge. By leveraging relational structure between data points, GCNs propagate information through the graph, enabling predictions for unlabeled nodes based on their neighbors. The core intuition is simple. If connected nodes tend to share similar properties, a phenomenon known as homophily, then a node's neighbors provide useful signal for predicting its label.

However, despite widespread adoption of GCNs, practitioners often lack clear guidance on when these methods actually help. The literature is rich with benchmark comparisons showing GCN improvements over baselines, but few studies systematically investigate the conditions under which GCNs succeed or fail. This gap between method development and practical deployment motivates our work.

We present a diagnostic study aimed at understanding not just whether GCNs work, but why they work and when they fail. Using the Amazon Computers dataset, a co-purchase graph where nodes represent products and edges indicate co-purchase relationships, we conduct systematic experiments to answer three questions. \textbf{First}, how does GCN performance change as labeled data becomes increasingly scarce? \textbf{Second}, what is the relative contribution of graph structure versus node features? \textbf{Third}, under what conditions does incorporating graph structure help or hurt?

Our experiments reveal several findings. GCNs provide the largest performance gains under extreme label scarcity, where they outperform feature-only baselines by a substantial margin. As more labels become available, this gap shrinks, suggesting that GCNs act as a safety net when supervision is limited. Through a feature ablation study, we find that GCNs can nearly match their original performance using structure alone, while feature-only baselines collapse to random performance. Most importantly, we find that neither homophily nor feature strength alone predicts when GCNs help. Instead, the interaction between these factors matters. GCNs help in three of four quadrants of the homophily-feature strength space but hurt when homophily is low and features are already strong. In this case, noisy neighbors corrupt predictions that would have been correct based on features alone.

These findings have practical implications. Before adopting a GCN, practitioners should assess the homophily of their graph and the predictive power of their features. If homophily is low and a simple baseline already performs well, adding graph structure may hurt rather than help.

\section{Related Work}\label{sec:related_work}

\textbf{Graph Neural Networks.} Graph Neural Networks have become the dominant approach for learning on graph-structured data. The key insight underlying most GNN architectures is message passing, where nodes iteratively aggregate information from their neighbors to update their representations \citep{Gilmer2017}. The Graph Convolutional Network introduced by \citet{Kipf2017} simplified earlier spectral approaches \citep{Bruna2014, Defferrard2016} by using a first-order approximation of spectral graph convolutions. Several variants address limitations of the vanilla GCN. Graph Attention Networks \citep{Velickovic2018} introduce attention mechanisms to weight neighbors differently, while GraphSAGE \citep{Hamilton2017} enables inductive learning through neighborhood sampling.

\textbf{Why We Choose Vanilla GCN.} We deliberately choose the vanilla two-layer GCN over more complex variants. Our dataset exhibits high homophily (77.7\%), where the attention mechanism in GAT provides limited benefit since most neighbors share the same label. Our setting is transductive, making GraphSAGE's inductive capabilities unnecessary. Most importantly, we aim to isolate the contribution of graph structure itself rather than architectural innovations. The vanilla GCN provides a clean baseline for this analysis.

\textbf{Semi-Supervised Learning on Graphs.} Semi-supervised node classification is one of the most studied problems in graph learning. Classical approaches include label propagation \citep{Zhu2003} and manifold regularization \citep{Belkin2006}. GCNs can be viewed as a neural generalization of these ideas, with neighborhood aggregation implicitly encouraging similar predictions for connected nodes. However, \citet{Shchur2018} identified pitfalls in GNN evaluation, showing sensitivity to data splits and hyperparameters. \citet{Huang2020} found that simple baselines can match GNN performance on some datasets when properly tuned.

\textbf{Homophily and Heterophily.} A key factor determining GNN effectiveness is homophily, the tendency of connected nodes to share similar properties. On homophilous graphs, neighborhood aggregation reinforces correct predictions. On heterophilous graphs, standard message passing can hurt performance. \citet{Zhu2020} systematically studied GNN performance across varying homophily levels and proposed architectural modifications for heterophilous graphs. Our work builds on these insights by examining how homophily interacts with feature quality to determine when GCNs help.

\textbf{Understanding GNN Behavior.} A growing body of work aims to understand why GNNs work. \citet{NT2019} showed that GCNs perform Laplacian smoothing, which can lead to oversmoothing in deep networks. Several studies have examined features versus structure. \citet{Huang2020} found that structure sometimes provides little benefit beyond features. \citet{Ma2022} proposed a framework analyzing when graph structure helps. Our quadrant analysis provides an empirical complement to such theoretical frameworks.

\section{Method}\label{sec:method}

\subsection{Problem Setting}

We study transductive semi-supervised node classification on a graph $G = (V, E)$ with $n = |V|$ nodes. Each node $v \in V$ has a feature vector $\mathbf{x}_v \in \mathbb{R}^d$ and a class label $y_v \in \{1, \ldots, C\}$. Node features are collected in a matrix $X \in \mathbb{R}^{n \times d}$, and graph structure is encoded in an adjacency matrix $A \in \{0, 1\}^{n \times n}$.

In the transductive setting, the entire graph structure and all node features are available during training, but labels are only observed for a subset of nodes $V_L \subset V$. The goal is to predict labels for the remaining unlabeled nodes $V_U = V \setminus V_L$.

To simulate varying degrees of label scarcity, we introduce a masking rate $m \in \{0\%, 50\%, 90\%\}$. A masking rate of $m$ means that $m$\% of training labels are hidden, leaving only $(100-m)$\% available for supervision. At 90\% masking, only 10\% of training nodes have observed labels.

\subsection{Graph Convolutional Network}

We use a standard two-layer GCN following \citet{Kipf2017}. The model consists of two graph convolutional layers with a ReLU activation between them.
\begin{align}
    H^{(1)} &= \text{ReLU}\left(\hat{A} X W^{(0)}\right) \\
    Z &= \hat{A} H^{(1)} W^{(1)}
\end{align}
Here $\hat{A} = \tilde{D}^{-1/2}\tilde{A}\tilde{D}^{-1/2}$ is the symmetrically normalized adjacency matrix with self-loops, $W^{(0)} \in \mathbb{R}^{d \times h}$ and $W^{(1)} \in \mathbb{R}^{h \times C}$ are learnable weight matrices, $h$ is the hidden dimension, and $C$ is the number of classes. The output $Z \in \mathbb{R}^{n \times C}$ contains logits for each node. We apply log-softmax and train using negative log-likelihood loss on labeled nodes.

The key property of this architecture is neighborhood aggregation. Each layer aggregates features from a node's immediate neighbors and itself via self-loops. After two layers, each node's representation incorporates information from its two-hop neighborhood. On homophilous graphs where neighbors tend to share labels, this aggregation reinforces correct class signal.

\subsection{Baseline Models}

We compare GCN against two feature-only baselines that ignore graph structure.

\textbf{Logistic Regression (LR).} We train a multinomial logistic regression classifier on node features with z-score normalization and tune regularization strength via cross-validation. LR provides a strong linear baseline measuring the predictive power of features alone.

\textbf{Support Vector Machine (SVM).} We train a linear SVM with class-balanced weighting and tune regularization via grid search optimizing macro-F1.

Both baselines operate solely on node features $X$ and ignore the adjacency matrix $A$. Any performance gap between GCN and these baselines can be attributed to graph structure. We use LR as our primary comparison because it consistently outperforms SVM on this dataset and provides calibrated probability estimates that allow us to interpret its performance as a measure of feature discriminability.

\subsection{Feature Ablation}

To isolate the contribution of graph structure versus node features, we conduct a feature ablation study by replacing original bag-of-words features with random Gaussian noise $\mathbf{x}_v \sim \mathcal{N}(0, I_d)$ for all nodes. The random features have the same dimensionality but carry no information about node labels. If a model relies primarily on features, its performance should collapse to near-random. If it relies primarily on structure, performance should remain stable.

\subsection{Homophily Measurement}

We quantify structural properties using edge homophily, defined as the fraction of edges connecting same-class nodes.
\begin{equation}
    h = \frac{|\{(u, v) \in E : y_u = y_v\}|}{|E|}
\end{equation}
A value of $h = 1$ indicates perfect homophily where all edges connect same-class nodes. High homophily ($h > 0.7$) suggests neighborhood aggregation should be beneficial. We also compute per-class homophily to understand variation across classes.

\subsection{Evaluation Protocol}

We split data into 80\% training and 20\% test using stratified sampling. For each masking rate, we further split training into sub-train and validation sets. We train for up to 200 epochs with early stopping based on validation macro-F1 using patience of 10 epochs.

We report macro-F1 as our primary metric, which computes the unweighted average of per-class F1 scores. Macro-F1 gives equal weight to each class regardless of size, making it more appropriate than accuracy for imbalanced datasets. To quantify GCN benefit, we compute $\Delta\text{F1} = \text{F1}_{\text{GCN}} - \text{F1}_{\text{LR}}$. Positive values indicate GCN outperforms the baseline.

\section{Experiments and Results}\label{sec:results}

\subsection{Dataset and Setup}

We evaluate on the Amazon Computers dataset \citep{Shchur2018, McAuley2015}, a co-purchase graph derived from Amazon's product recommendations. The graph contains 13,752 nodes representing products and 491,722 edges indicating co-purchase relationships. Each node has a 767-dimensional bag-of-words feature vector derived from product reviews, and one of 10 product category labels. The graph exhibits high homophily with 77.7\% of edges connecting same-class nodes, meaning customers who buy one product in a category tend to also buy other products in that same category. The complete neighbor distribution showing how each class connects to others is provided in Table~\ref{tab:neighbor_matrix} in the Appendix.

We use an 80/20 stratified train/test split. To simulate label scarcity, we mask 90\%, 50\%, or 0\% of training labels, representing scenarios where only 10\%, 50\%, or 100\% of training nodes have observed labels. All models use class-weighted loss to handle class imbalance, as the largest class contains 37.5\% of nodes while the smallest contains only 2.1\%. We report macro-F1 averaged across classes, which gives equal weight to each class regardless of size. Hyperparameter settings for all models are detailed in Table~\ref{tab:hyperparam} in the Appendix.

\subsection{GCN Helps Most Under Label Scarcity}

\begin{figure}[h!]
\centering
\includegraphics[width=\columnwidth]{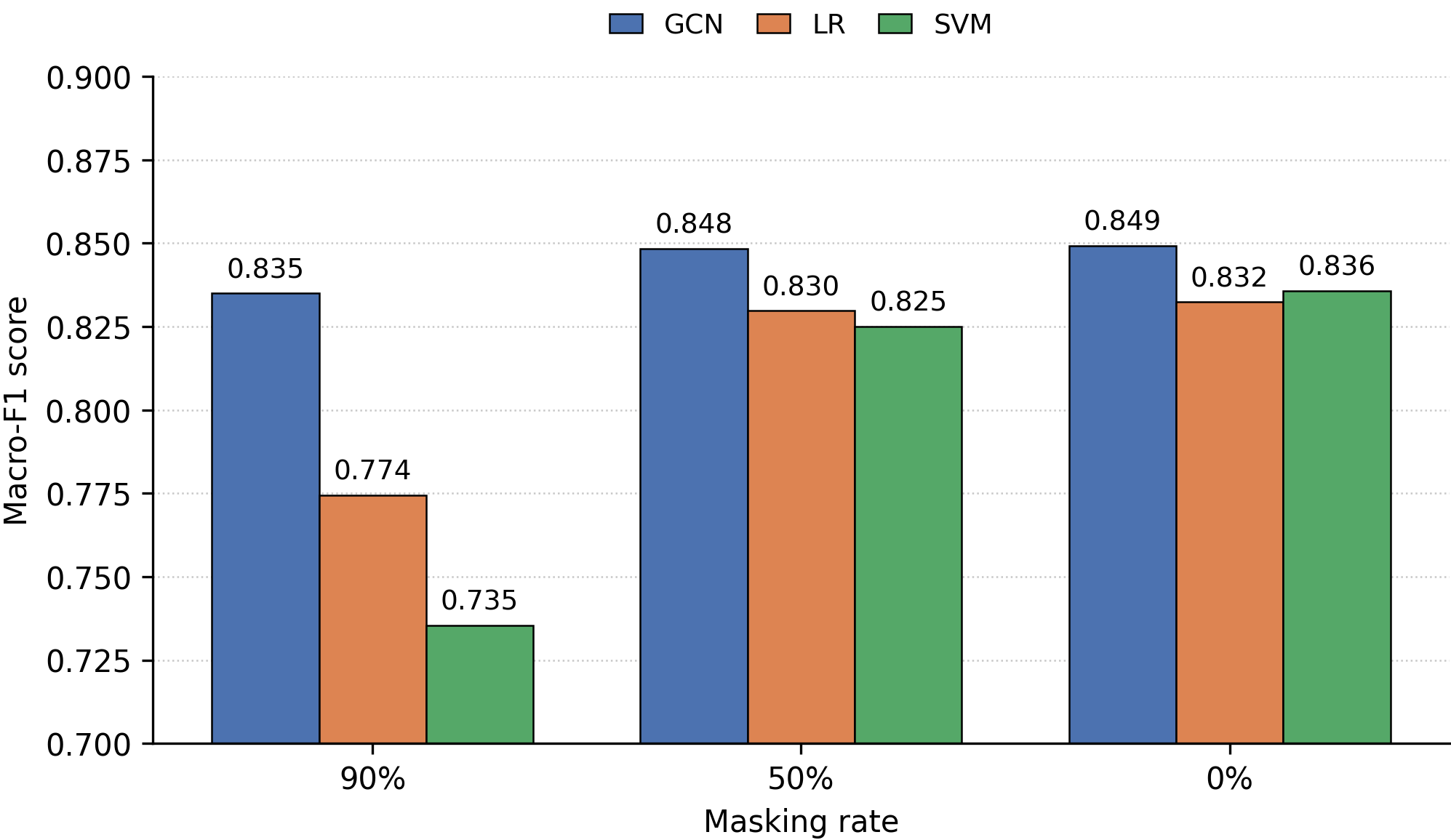}
\caption{Model performance across masking rates. GCN outperforms LR and SVM at all levels, with the largest gap at 90\% masking where GCN achieves 0.835 compared to LR's 0.775.}
\label{fig:masking_performance}
\end{figure}

We first examine how model performance changes as labeled data becomes increasingly scarce. Figure~\ref{fig:masking_performance} shows macro-F1 scores for GCN, Logistic Regression (LR), and SVM across the three masking conditions.

GCN consistently outperforms feature-only baselines across all conditions, but the performance gap varies substantially with label availability. At 90\% masking, where only 10\% of training labels are available, GCN achieves 0.835 macro-F1 compared to 0.775 for LR and 0.735 for SVM. This represents a 6.0 percentage point improvement over LR. As more labels become available, the gap shrinks. At 50\% masking, GCN leads LR by 2.1 points (0.851 vs 0.830). At 0\% masking with full supervision, the gap narrows further to 2.0 points (0.853 vs 0.833). Complete per-class breakdowns across all masking levels are provided in Table~\ref{tab:full_per_class} in the Appendix.

This pattern suggests that GCN acts as a safety net when supervision is limited. Under extreme label scarcity, feature-only models have insufficient training signal to learn discriminative patterns. GCN compensates by leveraging graph structure to propagate information from the few labeled nodes to their unlabeled neighbors. When a node's label is unknown but most of its neighbors share a common label, GCN can infer the correct class through neighborhood aggregation. As labels become abundant, this structural advantage diminishes because features alone become sufficient.

\subsection{Structure Alone Drives GCN Performance}

\begin{figure}[h!]
\centering
\includegraphics[width=\columnwidth]{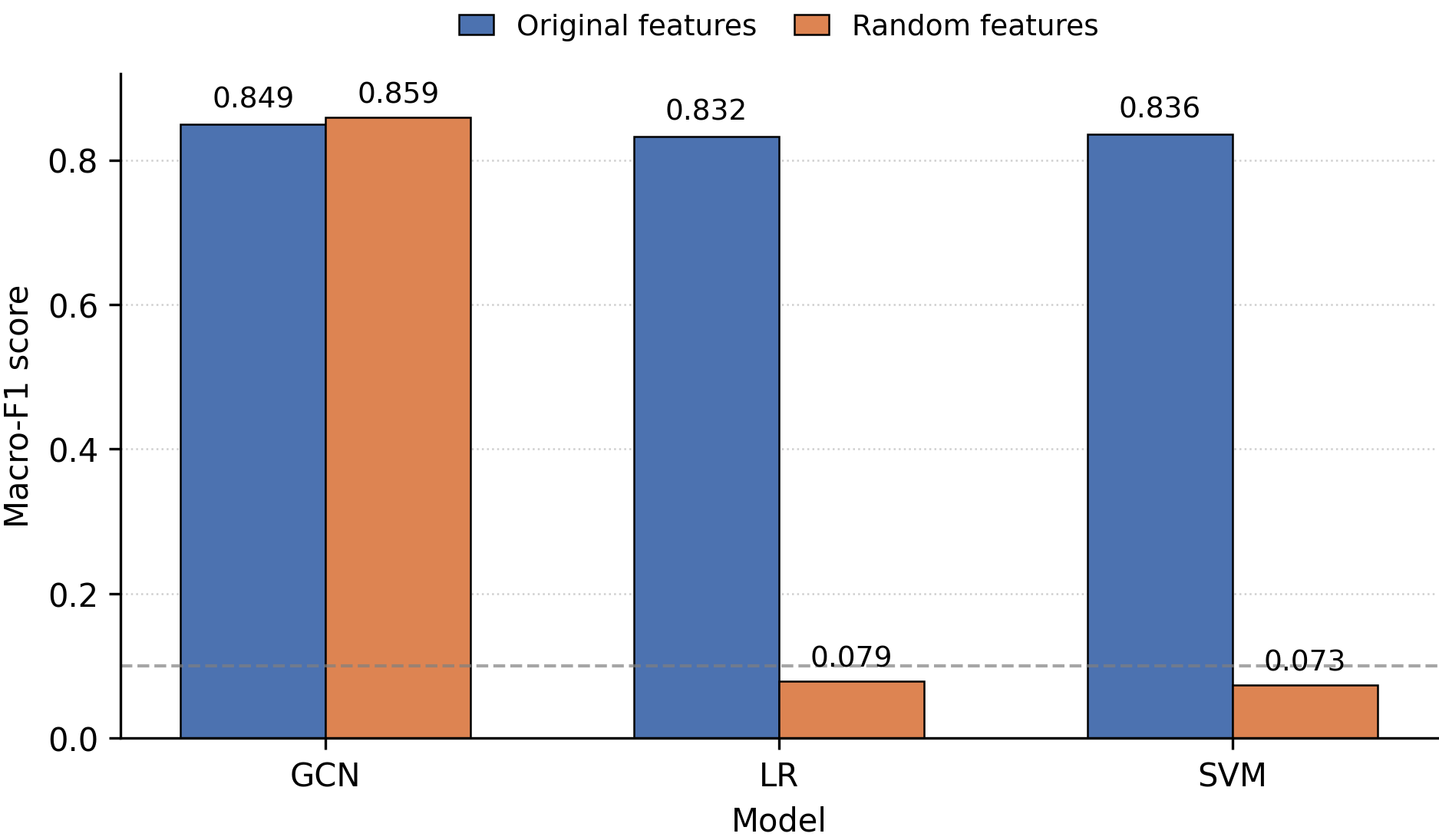}
\caption{Feature ablation at 0\% masking. With random features, LR and SVM collapse to near-random performance while GCN retains full performance, demonstrating that structure alone drives GCN predictions on this dataset.}
\label{fig:ablation}
\end{figure}

To understand whether GCN relies primarily on features or structure, we conducted a feature ablation study. We replaced all node features with random Gaussian noise, removing any information about node labels from the feature vectors. If a model relies on features, its performance should collapse. If it relies on structure, performance should remain stable. Figure~\ref{fig:ablation} shows results at 0\% masking.

The results reveal a stark contrast between model types. LR collapses from 0.833 to 0.079 F1, retaining only 9.5\% of its original performance. SVM similarly collapses from 0.836 to 0.073, retaining only 8.7\%. Both models perform near random guessing (10\% for 10 classes), confirming that they rely entirely on features. In contrast, GCN actually improves slightly from 0.853 to 0.859, retaining 100.7\% of its performance with random features.

This finding holds across all masking levels. As shown in Table~\ref{tab:ablation_all} in the Appendix, GCN retains between 97.6\% and 100.7\% of its performance across conditions when features are replaced with noise. The slight performance drop at 90\% masking (97.6\%) compared to the slight improvement at 0\% masking (100.7\%) suggests that under extreme label scarcity, the original features may provide a small amount of useful signal that complements structure.

Why can GCN succeed without meaningful features? The answer lies in the high homophily of this dataset. With 77.7\% of edges connecting same-class nodes, most of a node's neighbors share its label. GCN's neighborhood aggregation effectively performs a weighted vote among neighbors. When roughly 80\% of neighbors belong to the correct class, this vote is highly informative regardless of the feature vectors. The model learns to trust the neighborhood consensus, making features redundant. This also explains why removing the original features can slightly improve performance. The bag-of-words features may introduce noise that conflicts with the cleaner structural signal.

\subsection{Per-Class Performance Varies Widely}

While aggregate metrics show that GCN outperforms baselines on average, this masks significant heterogeneity at the class level. Table~\ref{tab:per_class} shows per-class F1 scores at 90\% masking, where the GCN advantage is largest.

\definecolor{MutedGreen}{RGB}{34,139,34}
\definecolor{MutedRed}{RGB}{139,0,0}

\begin{table}[h!]
\caption{Per-class F1 scores at 90\% masking. $\Delta$F1 shows GCN improvement over LR. Green indicates improvement, red indicates degradation.}
\label{tab:per_class}
\vskip 0.1in
\begin{center}
\begin{small}
\begin{tabular}{lccr}
\toprule
Class & GCN & LR & $\Delta$F1 \\
\midrule
0 & 0.889 & 0.661 & \textcolor{MutedGreen}{+0.228} \\
1 & 0.853 & 0.830 & \textcolor{MutedGreen}{+0.023} \\
2 & 0.938 & 0.877 & \textcolor{MutedGreen}{+0.061} \\
3 & 0.819 & 0.849 & \textcolor{MutedRed}{-0.030} \\
4 & 0.877 & 0.877 & +0.000 \\
5 & 0.946 & 0.830 & \textcolor{MutedGreen}{+0.116} \\
6 & 0.572 & 0.455 & \textcolor{MutedGreen}{+0.117} \\
7 & 0.869 & 0.849 & \textcolor{MutedGreen}{+0.020} \\
8 & 0.730 & 0.764 & \textcolor{MutedRed}{-0.034} \\
9 & 0.853 & 0.754 & \textcolor{MutedGreen}{+0.099} \\
\bottomrule
\end{tabular}
\end{small}
\end{center}
\vskip -0.1in
\end{table}

Most classes benefit substantially from GCN. Class 0 shows the largest improvement at +22.8 points, followed by Class 6 (+11.7), Class 5 (+11.6), and Class 9 (+9.9). These gains are practically significant and demonstrate the value of incorporating graph structure.

However, two classes show negative $\Delta$F1 where GCN actually hurts performance. Class 8 drops by 3.4 points and Class 3 by 3.0 points. For these classes, incorporating graph structure degrades predictions that would have been correct using features alone. This observation raises an important question. What distinguishes classes where GCN helps from those where it hurts?

\subsection{The Interaction Between Homophily and Feature Strength}

We tested two intuitive hypotheses about when GCN should help. First, high homophily should favor GCN because neighbors provide more informative signal when they share the same label. Second, weak baseline performance should favor GCN because there is more room for structural information to improve upon poor feature-based predictions.

Neither hypothesis holds in isolation. As shown in Figure~\ref{fig:scatter_appendix} in the Appendix, homophily versus $\Delta$F1 shows essentially no correlation. Some high-homophily classes benefit greatly while others show no improvement. Similarly, baseline F1 versus $\Delta$F1 shows only a weak negative trend with substantial scatter. Simple heuristics based on a single factor cannot reliably predict when GCN will help.

The answer lies in the interaction between these two factors. We divided classes into four quadrants based on median splits. Classes with homophily above 0.70 are considered high homophily. Classes where LR achieves F1 above 0.85 are considered to have strong features. The per-class homophily values used for this analysis are provided in Table~\ref{tab:class_homophily} in the Appendix. Figure~\ref{fig:quadrant} shows the resulting quadrant analysis.

\begin{figure}[h!]
\centering
\includegraphics[width=\columnwidth]{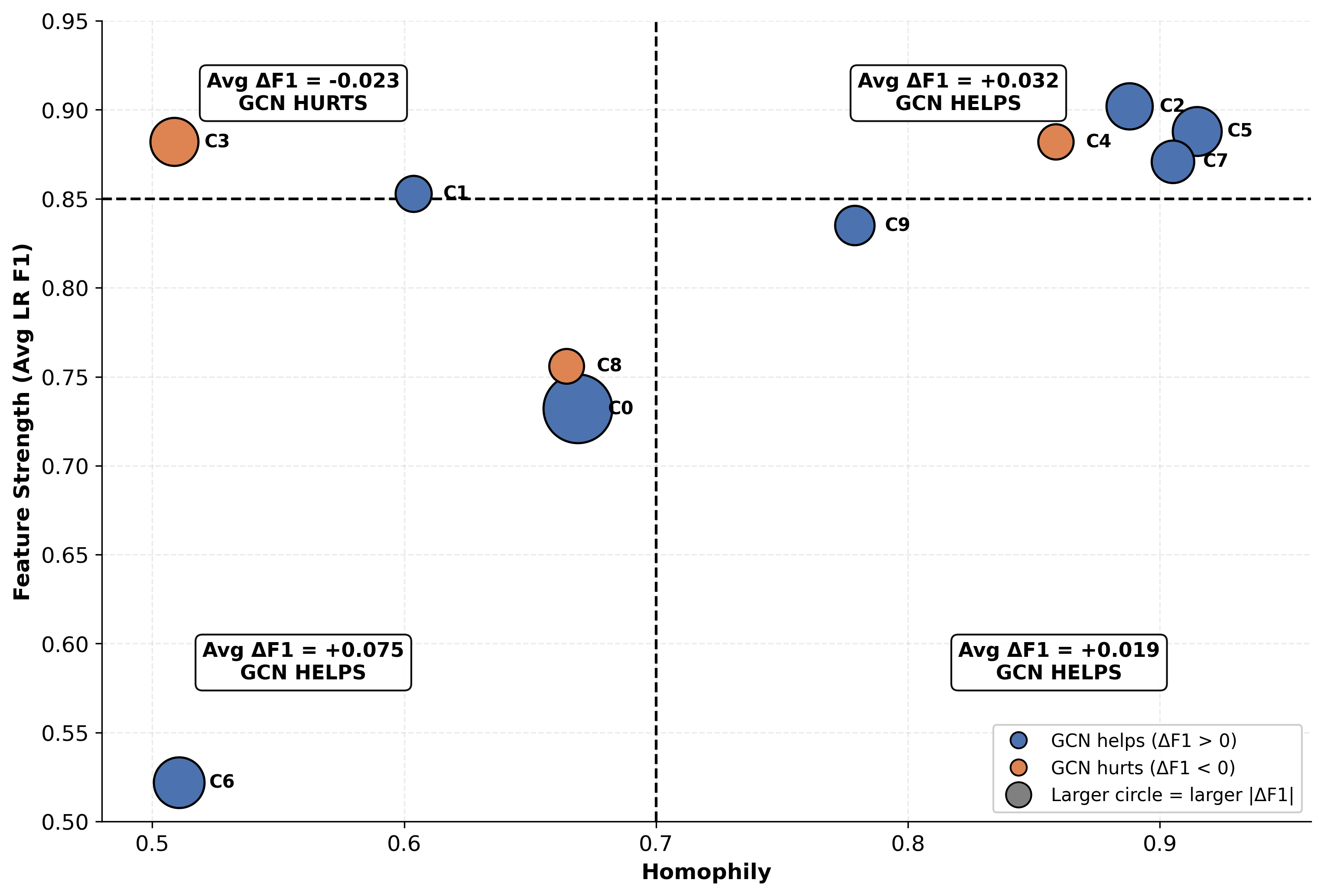}
\caption{Quadrant analysis of homophily and feature strength. Each point represents a class, sized by the magnitude of $\Delta$F1 and colored by whether GCN helps (green) or hurts (red). GCN hurts only in the top-left quadrant where low homophily meets strong features.}
\label{fig:quadrant}
\end{figure}

The pattern reveals a clear decision rule. In three of four quadrants, GCN helps.

In the top-right quadrant (high homophily, strong features), GCN provides modest improvements averaging +3.2 points. Clean neighbors reinforce already-good feature predictions. Classes 2, 4, 5, and 7 fall in this quadrant.

In the bottom-right quadrant (high homophily, weak features), GCN helps by +1.9 points on average. Structure compensates for weak features when neighbors are informative. Only Class 9 falls in this quadrant.

In the bottom-left quadrant (low homophily, weak features), GCN provides the largest improvements averaging +7.5 points. When features fail, even noisy structural signal is better than nothing. Classes 0, 6, and 8 fall here. Note that the quadrant analysis uses metrics averaged across all masking conditions, which is why Class 8 shows essentially no change ($\Delta$F1 = -0.003) despite the 3.4 point drop observed at 90\% masking in Table~\ref{tab:per_class}. The negative effect at high masking is offset by neutral or positive effects at lower masking rates.

GCN hurts only in the top-left quadrant (low homophily, strong features), where performance drops by 2.3 points on average. Classes 1 and 3 fall here. When homophily is low, neighbors provide noisy or misleading signal. When features are already strong, the baseline prediction is likely correct. Combining these conditions means GCN takes correct predictions and corrupts them by aggregating information from mixed-class neighbors. The detailed quadrant assignments with per-class metrics are provided in Table~\ref{tab:quadrant_classes} in the Appendix.

\subsection{Case Study of Low-Homophily Classes}

\begin{figure}[h!]
\centering
\includegraphics[width=\columnwidth]{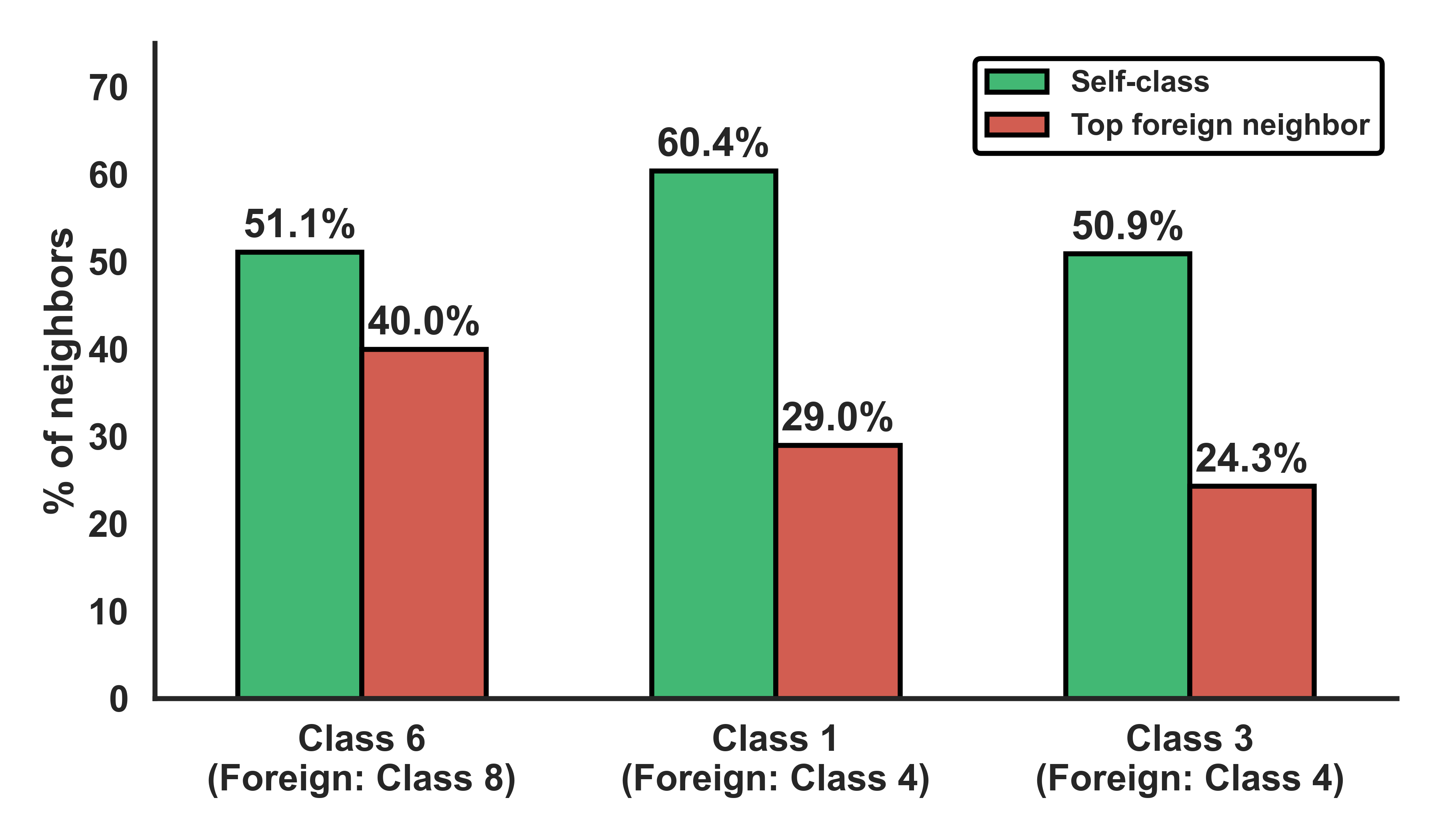}
\caption{Neighbor distributions for the three lowest-homophily classes. All have substantial foreign neighbor contamination, yet show different GCN outcomes depending on feature strength.}
\label{fig:low_homophily_classes}
\end{figure}

To understand these dynamics more concretely, we examine the three classes with lowest homophily. Figure~\ref{fig:low_homophily_classes} shows their neighbor distributions.

All three classes have substantial foreign neighbor contamination that creates challenging conditions for neighborhood aggregation. Class 6 has only 51.1\% same-class neighbors with a striking 40.0\% coming from Class 8, creating a near 50-50 split. Class 3 has 50.9\% same-class neighbors with 24.3\% from Class 4. Class 1 has 60.4\% same-class neighbors with 29.0\% from Class 4. Notably, Class 4 appears as a dominant foreign neighbor for multiple classes, suggesting it represents a broad product category that co-occurs with many others in purchasing patterns.

Despite similar homophily levels around 50-60\%, these classes show dramatically different GCN outcomes based on their feature strength. Class 6 has very weak features with LR achieving only 0.455 F1. GCN improves this substantially by +11.7 points to 0.572. Even though neighbors are noisy, the slight majority of same-class neighbors provides signal that beats the failing feature-based predictions. Class 1 has moderate features with LR F1 of 0.830. GCN still helps but modestly, improving by +2.3 points. Class 3 has strong features with LR F1 of 0.849. Here GCN hurts by -3.0 points. The noisy neighbors corrupt predictions that would have been largely correct using features alone.

The confusion matrix analysis in Table~\ref{tab:confusion_appendix} in the Appendix reveals the specific error patterns driving these outcomes. For Class 0, LR misclassifies 21.8\% of nodes as Class 4, while GCN reduces this dramatically to just 2.3\%. This explains the large +22.8 point improvement for Class 0. For Class 6, LR misclassifies 47.4\% as Class 8, while GCN reduces this to 19.2\%, cutting the error rate by more than half. In contrast, for Class 3, GCN increases misclassification to Class 4 from 0.9\% to 3.4\%, demonstrating how neighborhood aggregation can introduce errors when structural signal conflicts with correct feature-based predictions.

\subsection{Summary}

Our experiments reveal three key findings about GCN performance. First, GCN provides the largest benefits under label scarcity, acting as a safety net when supervision is limited. The 6.0 point improvement at 90\% masking shrinks to 2.0 points at 0\% masking. Second, on highly homophilous graphs, structure alone can drive accurate predictions, making features redundant. GCN retains 97-101\% of performance even with random features while baselines collapse. Third, GCN benefit depends critically on the interaction between homophily and feature strength rather than either factor alone. GCN helps in three of four quadrants and only hurts when low homophily meets strong features.

\section{Discussion}\label{sec:discussion}

\subsection{Practical Guidance}

Our findings suggest a simple diagnostic workflow for practitioners considering GCNs. Before adopting a graph-based method, compute edge homophily on your graph. If homophily exceeds 0.7, GCN is likely to help regardless of feature quality. Next, train a logistic regression baseline on features alone. If this baseline performs well (F1 above 0.85) and homophily is low, adding graph structure may hurt rather than help. The label budget also matters. GCN provides the largest benefit when labeled data is limited, so practitioners with abundant labels may find simpler methods sufficient.

Beyond aggregate metrics, examining per-class performance can reveal hidden problems. Some classes may be hurt by GCN even when overall performance improves. For underperforming classes, investigating neighbor distributions can reveal whether high foreign-class contamination is corrupting predictions.

\subsection{Limitations}

Our analysis focuses on a single dataset with high average homophily (77.7\%) and bag-of-words features. While we observed substantial per-class variation in homophily, validating our quadrant framework on datasets with lower global homophily would strengthen its generalizability.

We also studied only the vanilla two-layer GCN without architectural modifications. Methods designed for heterophilous graphs, or architectures with attention mechanisms like GAT, might behave differently in the low-homophily regime where attention could learn to downweight noisy neighbors.

Our feature ablation uses random Gaussian noise, representing complete feature removal. Intermediate ablations that degrade features gradually could reveal more nuanced interactions. Similarly, our quadrant analysis used median splits for thresholds, and a more principled approach might learn optimal thresholds from data or use continuous measures.

\subsection{Future Directions}

Replicating our diagnostic framework across datasets with varying homophily levels would test the generality of the quadrant pattern. Comparing vanilla GCN with GAT and GraphSAGE would reveal whether our findings are architecture-specific or general to neighborhood aggregation. Developing methods to automatically predict when GCN will help based on observable graph properties would provide actionable guidance without extensive experimentation. Testing whether architectural modifications such as edge dropout or attention mechanisms can mitigate the negative effects in the low-homophily strong-features regime remains an open question with practical significance.

\section{Conclusion}

Graph structure is not universally helpful for node classification. The benefit of GCNs depends critically on the interplay between graph homophily and feature quality. Our quadrant framework offers a simple diagnostic based on two measurable quantities that practitioners can compute before committing to a GNN architecture. We hope this work encourages more diagnostic studies that move beyond benchmark comparisons to understand the mechanisms underlying GNN performance.

\bibliography{ref}
\bibliographystyle{icml2025}

\newpage
\appendix
\onecolumn
\section{Additional Results}\label{sec:appendix}

\subsection{Feature Ablation Across Masking Levels}

Table~\ref{tab:ablation_all} shows GCN performance with original versus random features across all masking conditions. GCN retains 97-101\% of its performance even when features carry no information, confirming that structure alone drives predictions on this dataset.

\begin{table}[h]
\caption{GCN performance with original vs. random features.}
\label{tab:ablation_all}
\vskip 0.1in
\begin{center}
\begin{small}
\begin{tabular}{lccc}
\toprule
Masking & Original & Random & \% Retained \\
\midrule
90\% & 0.835 & 0.815 & 97.6\% \\
50\% & 0.851 & 0.857 & 100.7\% \\
0\% & 0.853 & 0.859 & 100.7\% \\
\bottomrule
\end{tabular}
\end{small}
\end{center}
\end{table}

\subsection{Single-Factor Correlations}

\begin{figure}[h]
\centering
\includegraphics[width=0.6\columnwidth]{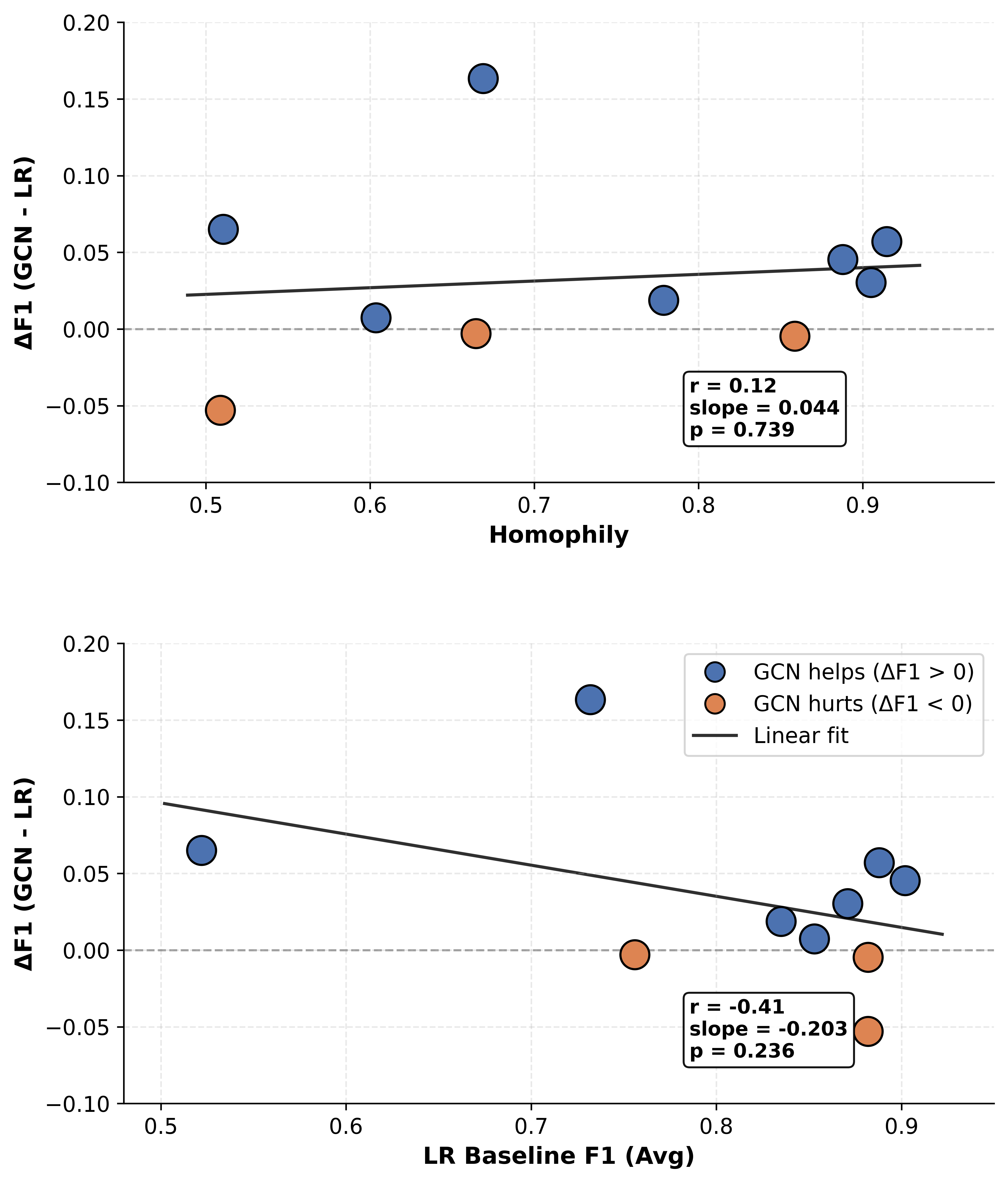}
\caption{Single-factor correlations fail to predict GCN benefit. Neither homophily (top) nor baseline strength (bottom) alone explains when GCN helps.}
\label{fig:scatter_appendix}
\end{figure}

Figure~\ref{fig:scatter_appendix} shows that neither homophily nor baseline strength alone predicts GCN benefit. The left plot shows homophily versus $\Delta$F1 with no clear correlation. The right plot shows LR baseline F1 versus $\Delta$F1 with only a weak negative trend and substantial scatter.

\subsection{Confusion Matrix Analysis}

Table~\ref{tab:confusion_appendix} shows misclassification rates for key class pairs at 90\% masking. GCN dramatically reduces Class 0 $\to$ Class 4 errors (2.3\% vs 21.8\%) and Class 6 $\to$ Class 8 errors (19.2\% vs 47.4\%), explaining the large improvements for these classes. However, GCN increases Class 3 $\to$ Class 4 errors (3.4\% vs 0.9\%), demonstrating how noisy neighbors can corrupt good feature-based predictions.

\begin{table}[h]
\caption{Top misclassification pairs at 90\% masking.}
\label{tab:confusion_appendix}
\vskip 0.1in
\begin{center}
\begin{small}
\begin{tabular}{lcc}
\toprule
Confusion Pair & GCN Error & LR Error \\
\midrule
Class 6 $\to$ Class 8 & 19.2\% & 47.4\% \\
Class 8 $\to$ Class 6 & 18.6\% & 9.7\% \\
Class 0 $\to$ Class 4 & 2.3\% & 21.8\% \\
Class 1 $\to$ Class 4 & 7.8\% & 7.5\% \\
Class 3 $\to$ Class 4 & 3.4\% & 0.9\% \\
\bottomrule
\end{tabular}
\end{small}
\end{center}
\end{table}

\subsection{Complete Per-Class Results}

Table~\ref{tab:full_per_class} provides per-class F1 scores across all masking conditions and models.

\begin{table}[h]
\caption{Complete per-class F1 scores across masking conditions.}
\label{tab:full_per_class}
\vskip 0.1in
\begin{center}
\begin{small}
\begin{tabular}{l|ccc|ccc|ccc}
\toprule
 & \multicolumn{3}{c|}{90\% Masking} & \multicolumn{3}{c|}{50\% Masking} & \multicolumn{3}{c}{0\% Masking} \\
Class & GCN & LR & SVM & GCN & LR & SVM & GCN & LR & SVM \\
\midrule
0 & .889 & .661 & .694 & .903 & .770 & .777 & .911 & .771 & .802 \\
1 & .853 & .830 & .788 & .869 & .866 & .854 & .870 & .865 & .873 \\
2 & .938 & .877 & .856 & .950 & .915 & .913 & .950 & .915 & .919 \\
3 & .819 & .849 & .785 & .834 & .884 & .875 & .815 & .913 & .881 \\
4 & .877 & .877 & .855 & .883 & .886 & .884 & .884 & .881 & .887 \\
5 & .946 & .830 & .787 & .961 & .920 & .908 & .961 & .916 & .909 \\
\rowcolor{red!10} 6 & .572 & .455 & .395 & .590 & .543 & .537 & .604 & .565 & .582 \\
7 & .869 & .849 & .767 & .903 & .885 & .873 & .918 & .879 & .856 \\
8 & .730 & .764 & .709 & .755 & .754 & .755 & .764 & .749 & .753 \\
9 & .853 & .754 & .717 & .863 & .874 & .875 & .855 & .878 & .894 \\
\midrule
Macro & .835 & .775 & .735 & .851 & .830 & .825 & .853 & .833 & .836 \\
\bottomrule
\end{tabular}
\end{small}
\end{center}
\end{table}

\subsection{Per-Class Homophily and Quadrant Assignments}

Table~\ref{tab:class_homophily} shows homophily values and top foreign neighbor for each class.

\begin{table}[h]
\caption{Per-class homophily and top foreign neighbor.}
\label{tab:class_homophily}
\vskip 0.1in
\begin{center}
\begin{small}
\begin{tabular}{lccc}
\toprule
Class & Homophily & Top Foreign & \% Foreign \\
\midrule
0 & 66.9\% & Class 3 & 12.2\% \\
1 & 60.4\% & Class 4 & 29.0\% \\
2 & 88.8\% & Class 4 & 3.4\% \\
3 & 50.9\% & Class 4 & 24.3\% \\
4 & 85.9\% & Class 1 & 6.9\% \\
5 & 91.5\% & Class 4 & 5.6\% \\
6 & 51.1\% & Class 8 & 40.0\% \\
7 & 90.5\% & Class 2 & 4.9\% \\
8 & 66.5\% & Class 4 & 13.9\% \\
9 & 77.9\% & Class 4 & 14.8\% \\
\midrule
Overall & 77.7\% & -- & -- \\
\bottomrule
\end{tabular}
\end{small}
\end{center}
\end{table}

Table~\ref{tab:quadrant_classes} shows quadrant assignments based on median splits of homophily (0.70) and LR F1 (0.85). Values are averaged across masking conditions.

\begin{table}[h]
\caption{Quadrant assignments with per-class metrics (averaged across masking).}
\label{tab:quadrant_classes}
\vskip 0.1in
\begin{center}
\begin{small}
\begin{tabular}{llccc}
\toprule
Quadrant & Class & Homophily & LR F1 & $\Delta$F1 \\
\midrule
\multirow{2}{*}{Low H, Strong F} & C1 & 0.604 & 0.853 & +0.007 \\
 & C3 & 0.509 & 0.882 & -0.053 \\
\midrule
\multirow{4}{*}{High H, Strong F} & C2 & 0.888 & 0.902 & +0.045 \\
 & C4 & 0.859 & 0.882 & -0.005 \\
 & C5 & 0.915 & 0.888 & +0.057 \\
 & C7 & 0.905 & 0.871 & +0.030 \\
\midrule
\multirow{3}{*}{Low H, Weak F} & C0 & 0.669 & 0.732 & +0.163 \\
 & C6 & 0.511 & 0.522 & +0.065 \\
 & C8 & 0.665 & 0.756 & -0.003 \\
\midrule
High H, Weak F & C9 & 0.779 & 0.835 & +0.019 \\
\bottomrule
\end{tabular}
\end{small}
\end{center}
\end{table}

\subsection{Neighbor Distribution Matrix}

Table~\ref{tab:neighbor_matrix} shows the full neighbor distribution. Each row indicates what percentage of a class's neighbors belong to each class. Diagonal entries (bold) show self-class homophily. Asterisks mark high foreign concentration ($\geq$15\%).

\begin{table}[h]
\caption{Neighbor distribution matrix (\%).}
\label{tab:neighbor_matrix}
\vskip 0.1in
\begin{center}
\begin{small}
\begin{tabular}{l|cccccccccc}
\toprule
 & C0 & C1 & C2 & C3 & C4 & C5 & C6 & C7 & C8 & C9 \\
\midrule
C0 & \textbf{66.9} & 2.4 & 6.4 & 12.2 & 7.5 & 0.0 & 0.4 & 0.6 & 3.5 & 0.1 \\
C1 & 0.4 & \textbf{60.4} & 1.0 & 1.8 & *29.0 & 0.0 & 0.8 & 0.2 & 6.3 & 0.1 \\
C2 & 1.7 & 1.4 & \textbf{88.8} & 1.1 & 3.4 & 0.1 & 0.2 & 2.0 & 1.2 & 0.0 \\
C3 & 8.1 & 6.4 & 2.8 & \textbf{50.9} & *24.3 & 0.3 & 0.6 & 0.3 & 6.0 & 0.3 \\
C4 & 0.3 & 6.9 & 0.6 & 1.6 & \textbf{85.9} & 0.2 & 0.3 & 0.1 & 3.8 & 0.5 \\
C5 & 0.0 & 0.3 & 0.3 & 0.5 & 5.6 & \textbf{91.5} & 0.3 & 0.5 & 0.8 & 0.4 \\
C6 & 0.3 & 2.9 & 0.5 & 0.5 & 3.9 & 0.2 & \textbf{51.1} & 0.6 & *40.0 & 0.1 \\
C7 & 0.4 & 0.6 & 4.9 & 0.3 & 1.1 & 0.3 & 0.6 & \textbf{90.5} & 1.3 & 0.0 \\
C8 & 0.6 & 5.5 & 0.7 & 1.5 & 13.9 & 0.1 & 10.4 & 0.3 & \textbf{66.5} & 0.5 \\
C9 & 0.2 & 1.1 & 0.1 & 0.6 & 14.8 & 0.4 & 0.2 & 0.1 & 4.6 & \textbf{77.9} \\
\bottomrule
\end{tabular}
\end{small}
\end{center}
\end{table}

\subsection{Hyperparameter Selection}

Table~\ref{tab:hyperparam} shows hyperparameter search spaces and selected values. GCN performed best with no weight decay, while both LR and SVM selected small regularization values indicating strong regularization helps on this high-dimensional feature space.

\begin{table}[h]
\caption{Hyperparameter search space and selected values.}
\label{tab:hyperparam}
\vskip 0.1in
\begin{center}
\begin{small}
\begin{tabular}{llll}
\toprule
Model & Parameter & Search Space & Selected \\
\midrule
\multirow{4}{*}{GCN} 
 & hidden\_channels & \{32, 64, 128\} & 64 \\
 & dropout\_rate & \{0.2, 0.3, 0.5\} & 0.5 \\
 & learning\_rate & \{0.001, 0.01\} & 0.01 \\
 & weight\_decay & \{5e-4, 1e-4, 1e-5, 0\} & 0 \\
\midrule
LR 
 & C & \{0.001, 0.01, 0.1, 1, 10, 100, 1000\} & 0.01 \\
\midrule
SVM 
 & C & \{0.0001, 0.001, 0.01, 0.1, 1, 10, 100\} & 0.001 \\
\bottomrule
\end{tabular}
\end{small}
\end{center}
\end{table}

\end{document}